\pdfoutput=1
\documentclass[11pt]{article}
\usepackage{overpic}
\usepackage{float}

\usepackage[a-1b]{pdfx}

\usepackage{framed}
\usepackage{lmodern}
\usepackage{mdwlist}
\usepackage{siunitx}
\usepackage{latexsym}
\usepackage{colortbl}
\usepackage{xcolor}
\usepackage{nicefrac}
\usepackage{booktabs}
\usepackage{fnpct}
\usepackage{amsfonts}
\usepackage[T1]{fontenc}
\usepackage{bold-extra}
\usepackage{amsmath}
\usepackage{amssymb}
\usepackage{bm}
\usepackage{graphicx}
\usepackage{mathtools}
\usepackage{microtype}
\usepackage{multirow}
\usepackage{multicol}
\usepackage{xpatch}
\usepackage{latexsym,comment}
\usepackage[normalem]{ulem}

\newcommand*{\missingreference}{{\Huge \colorbox{red}{?reference?}}}
\newcommand*{\missingcitation}{{\Huge \colorbox{red}{?citation?}}}

\makeatletter
\xpatchcmd{\@setref}{\bfseries}{\missingreference}{}{}
\def\@citex[#1]#2{\leavevmode
    \let\@citea\@empty
    \@cite{\@for\@citeb:=#2\do
        {\@citea\def\@citea{,\penalty\@m\ }%
            \edef\@citeb{\expandafter\@firstofone\@citeb\@empty}%
            \if@filesw\immediate\write\@auxout{\string\citation{\@citeb}}\fi
            \@ifundefined{b@\@citeb}{\hbox{\reset@font\missingcitation}%
                \G@refundefinedtrue
                \@latex@warning
                {Citation `\@citeb' on page \thepage \space undefined}}%
            {\@cite@ofmt{\csname b@\@citeb\endcsname}}}}{#1}}
\makeatother

\newcommand{\ui}[0]{UI\xspace}
\newcommand{\irt}[0]{IRT\xspace}
\newcommand{\mmlu}[0]{MMLU\xspace}
\newcommand{\mrc}[0]{MRC\xspace}
\newcommand{\lm}[0]{LM\xspace}
\newcommand{\nli}[0]{NLI\xspace}
\newcommand{\sat}[0]{SAT\xspace}
\newcommand{\mm}[0]{LLM\xspace}
\newcommand{\mcqa}{MCQA\xspace}
\newcommand{\emcqa}{E-MCQA\xspace}
\newcommand{\mcq}{MCQ\xspace}
\newcommand{\success}{SR\xspace}

\newcommand{\gem}[1]{\mbox{\textsc{gem}}}
\newcommand{\abr}[1]{\textsc{#1}\xspace}

\newcommand{\hidetext}[1]{}
\newcommand{\ignore}[1]{}
\newif\ifcomment
\commenttrue

\ifcomment
    \newcommand{\pinaforecomment}[3]{\colorbox{#1}{\parbox{.8\linewidth}{#2: #3}}}

    \newcommand{\prtodo}[1]{\pinaforecomment{lightblue}{pr}{#1}}
    \newcommand{\prtodoi}[1]{\pinaforecomment{lightblue}{pr}{#1}}
\else
    \newcommand{\pinaforecomment}[3]{}
    \newcommand{\prtodo}[1]{}
    \newcommand{\prtodoi}[1]{}
\fi

\newcommand{\smallurl}[1]{ \begin{tiny}\url{#1}\end{tiny}}

\definecolor{lightblue}{HTML}{3cc7ea}
\definecolor{CUgold}{HTML}{CFB87C}
\definecolor{grey}{rgb}{0.95,0.95,0.95}
\definecolor{ceil}{rgb}{0.57, 0.63, 0.81}
\definecolor{UMDred}{HTML}{ed1c24}
\definecolor{UMDyellow}{HTML}{ffc20e}

\newcommand{\qa}[0]{\abr{qa}}

\newcommand{\nlp}[0]{\abr{nlp}}

\usepackage[]{acl2023}

\usepackage{xspace}
\usepackage{xcolor}
\usepackage[utf8]{inputenc}
\usepackage{pgfplots}
\usepackage{dsfont}
\DeclareUnicodeCharacter{2212}{−}
\usepgfplotslibrary{groupplots,dateplot}
\usetikzlibrary{patterns,shapes.arrows}
\pgfplotsset{compat=newest}

\usepackage{tikzscale}
\usepackage{relsize}
\usepackage{amsmath,amssymb}
\newcommand{\probP}{\text{I\kern-0.15em P}}

\usepackage{multirow, colortbl}

\usepackage{tabularx,booktabs}
\usepackage{makecell}

\usepackage[normalem]{ulem}
\useunder{\uline}{\ul}{}

\usepackage{graphicx}

\definecolor{ablation6}{HTML}{fcefed}
\definecolor{ablation_tie}{HTML}{fce3e1}

\definecolor{ablation5}{HTML}{fcd8d4}
\definecolor{ablation4}{HTML}{FBC3BC}
\definecolor{ablation3}{HTML}{F7A399}
\definecolor{ablation2}{HTML}{F38375}
\definecolor{ablation1}{HTML}{EF6351}

\newcommand{\inlinecode}[1]{%
    \begin{tikzpicture}[baseline=0ex]%
         \node[anchor=base,%
         text height=0.7em,%
         text depth=0.7ex,%
         inner ysep=0pt,%
         draw=lightgray!50,%
         fill=lightgray!50,%
         rounded corners=2pt] at (0,0) {\footnotesize\texttt{#1}};%
    \end{tikzpicture}%
}

\usepackage[]{algpseudocode}
\usepackage[]{algorithm}
\usepackage{float}
\algtext*{EndFor}%
\algtext*{EndProcedure}%

\usepackage{booktabs}
\usepackage[normalem]{ulem}
\useunder{\uline}{\ul}{}

\usepackage{times}
\usepackage{latexsym}
\usepackage{adjustbox}

\usepackage[T1]{fontenc}
\usepackage{amsmath}
\usepackage{amssymb}
\usepackage{booktabs}
\usepackage{tikzscale}
\usepackage{amsmath}

\usepackage{multirow, colortbl}

\usepackage{tabularx,booktabs}
\usepackage{makecell}
\usepackage{multirow}
\usepackage{scalerel,xparse}
\usepackage{cleveref}
\usepackage{microtype}
\usepackage[most]{tcolorbox}

\usepackage{enumitem}
\crefformat{section}{\S#2#1#3}
\crefformat{subsection}{\S#2#1#3}
\crefformat{subsubsection}{\S#2#1#3}

\definecolor{bggray}{rgb}{0.95, 0.95, 0.95}
\usepackage[%
    framemethod=tikz,
    skipbelow=\topskip,
    skipabove=\topskip
]{mdframed}
\mdfsetup{%
    leftmargin=0pt,
    rightmargin=0pt,
    backgroundcolor=bggray,
    middlelinecolor=black,
    roundcorner=3
}

\definecolor{SkyBlue}{rgb}{0.53, 0.81, 0.92}

\newtcolorbox[list inside=prompt,auto counter,number within=section]{prompt}[1][]{
    colbacktitle=black!60,
    fonttitle=\small,
    coltitle=white,
    fontupper=\footnotesize,
    boxsep=3pt,
    left=0pt,
    right=0pt,
    top=0pt,
    bottom=0pt,
    boxrule=1pt,
    #1
}

\definecolor{UMDred}{HTML}{ed1c24}

\definecolor{yellowcolor}{HTML}{ffc20e}
\definecolor{redcolor}{HTML}{e99999}
\definecolor{orangecolor}{HTML}{f6b26b}
\definecolor{yellowcolor}{HTML}{ffd966}
\definecolor{bluecolor}{HTML}{a0c5e8}
\definecolor{purplecolor}{HTML}{d9d2e9}

\title{Which of These Best Describes Multiple Choice Evaluation with LLMs?\\A) Forced B) Flawed C) Fixable \colorbox{yellow}{D) All of the Above}}

\author{Nishant Balepur \\
  University of Maryland\\
  \texttt{nbalepur@umd.edu} \\\And
  Rachel Rudinger \\
  University of Maryland \\
  \texttt{rudinger@umd.edu}
  \\\And
  Jordan Boyd-Graber \\
  University of Maryland \\
  \texttt{jbg@umiacs.umd.edu}}

\begin{document}
\maketitle

\begin{abstract} {
Multiple choice question answering (\mcqa{}) is popular for \mm{} evaluation due to its simplicity and human-like testing, but we argue for its reform.
We first reveal flaws in~\mcqa{}'s format, as it struggles to: 1) test generation/subjectivity; 2) match \mm{} use cases; and 3) fully~test knowledge.
We instead advocate for generative formats based on human testing---where \mm{}s construct
and explain answers---better~capturing user needs and knowledge while remaining easy to score.
We then show even when \mcqa{} is a useful format, its datasets suffer from: leakage; unanswerability; shortcuts; and saturation.
In each issue, we give fixes from education, like rubrics to guide \mcq{} writing; scoring methods to bridle guessing; and Item Response Theory to build harder \mcq{}s.
Lastly, we discuss \mm{} errors in \mcqa{}---robustness, biases, and unfaithful explanations---showing how our prior solutions better measure or address these issues.
While we do not need to desert \mcqa{}, we encourage more efforts in refining the task based on educational testing, advancing evaluations.
}
\end{abstract}

\section{Questioning Multiple Choice Questions}

Multiple choice question answering (\mcqa) is~the standard for large language model (\mm{}) evaluations, prized for simplicity and
similarity to human testing~\cite{robinson2023leveraging}.
When designing new benchmarks, \mcqa seems easy to~implement
\cite{guo2023evaluating}, and when selecting new \mm{}s to use, \mcqa
leaderboards inform our decisions~\cite{open-llm-leaderboard-v2}.
If you want to build a popular dataset, prove your \mm{} is smart,
or even publish a position paper, it is hard to avoid~\mcqa.

\begin{figure}
    \centering
    \begin{overpic}[width=\linewidth]{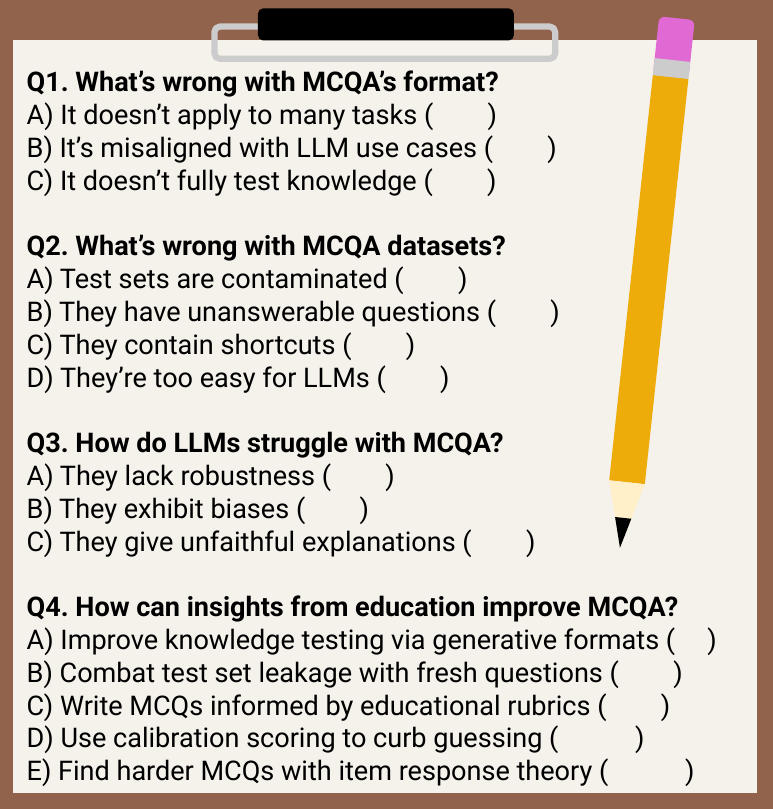}

    \put(53.6,84.5){\scalebox{0.8}{\cref{subsection:best_answer}}}
    \put(60.9,80.4){\scalebox{0.8}{\cref{subsection:use_cases}}}
    \put(53.5,76.3){\scalebox{0.8}{\cref{subsection:testing_what}}}

    \put(49.9,64.2){\scalebox{0.8}{\cref{subsection:test_set_leakage}}}   
    \put(61.3,60.1){\scalebox{0.8}{\cref{subsection:quality}}} 
    \put(43.4,56.0){\scalebox{0.8}{\cref{subsection:artifacts}}} 
    \put(47.6,51.9){\scalebox{0.8}{\cref{subsection:saturation}}} 

    \put(40.9,39.8){\scalebox{0.8}{\cref{subsection:robustness}}}  
    \put(37.7,35.7){\scalebox{0.8}{\cref{subsection:bias}}}  
    \put(58.2,31.6){\scalebox{0.8}{\cref{subsection:explanations}}} 

    \put(83.5,19.5){\scalebox{0.8}{\cref{section:fixing_format}}}
    \put(76.5,15.4){\scalebox{0.8}{\cref{subsection:test_set_leakage}}} 
    \put(74.9,11.3){\scalebox{0.8}{\cref{subsection:quality}}} 
    \put(68.9,7.3){\scalebox{0.8}{\cref{subsubsection:shortcut_scoring}}} 
    \put(75.2,3.2){\scalebox{0.8}{\cref{subsection:irt}}} 
    
    \end{overpic}
    \vspace{-3ex}
    \caption{Overview of this paper. We show many problems in formats (\cref{section:format}), datasets (\cref{section:dataset}), and LLMs (\cref{section:models}) when using \mcqa. Along the way, we propose solutions and ideas for future work, drawing from educational testing.}
    \label{fig:intro}
    \vspace{-1ex}
\end{figure}

Standardized testing groups have long explored ways to
better use \mcqa for student testing~\cite{angoff1971college}.
But despite years of use in \abr{nlp}~\cite{turney2003combining}, few have asked: 1) should \mcqa be a standard model evaluation format; and 2) are its datasets well-designed?
This position paper argues: \textbf{Evaluating \mm{}s with \mcqa has flaws in both its inherent format and dataset construction.}
We state our position in three points (Fig~\ref{fig:intro}).

We first argue \mcqa is not an ideal standardized format for \mm{} evaluations, showing its goal of ``pick the
best answer'' cannot optimally test generation or subjectivity
(\cref{subsection:best_answer}), misaligns with \mm{} use cases
(\cref{subsection:use_cases}), and poorly tests knowledge
(\cref{subsection:testing_what}).
Drawing from education, we advocate two \textit{generative}
improvements to \mcqa's format for future exploration: 1) providing
short, constructed-response answers without using choices
(\cref{subsection:constructed_response}); and 2) evaluating
explanations for model answers (\cref{subsection:justified_mcqa}).
These formats capture generation or subjectivity, match \mm{} use cases,
and improve knowledge testing, all while mostly preserving \mcqa's
simple scoring.

Next, we argue even when \mcqa is a useful format, its datasets suffer from: dataset leakage
(\cref{subsection:test_set_leakage}), unanswerable \mcq{}s~(\cref{subsection:quality}), shortcuts (\cref{subsection:artifacts}),
and saturation (\cref{subsection:saturation}), degrading \mcqa's utility.
To enhance \abr{nlp} dataset design, we offer solutions for each issue based on best practices in human
testing, like rubrics to flag \mcq errors (\cref{subsection:quality}),
metrics~to curb guessing from shortcuts
(\cref{subsubsection:shortcut_scoring}), and Item Response Theory
\cite{baker2001basics} to cull shoddy \mcq{}s and make the ones left
more challenging~(\cref{subsection:irt}).

Lastly, we show many errors of \mm{}s in \mcqa directly relate to
\mcqa's flaws (\cref{section:models}).
These issues, like brittleness to perturbations
(\cref{subsection:robustness}), bias toward certain options, cultures,
and languages (\cref{subsection:bias}), and generating
unfaithful explanations (\cref{subsection:explanations}), can all be
better measured or addressed with our proposed improvements to \mcqa's
format and datasets.

Many promising improvements to \mcqa draw from education, a field dedicated to effective assessment
\cite{haladyna2002review}, but these practices are rarely used in \abr{nlp}.
Adopting them demands more effort---\mcqa{} is popular
as it seems simple---but this effort is worth it to improve
evaluations.
To encourage researchers to take on these challenges, we conclude with
guidelines for designing meaningful evaluations whether or not you use
\mcqa{}~(\cref{section:conclusion}).

\section{Background: A Brief History of \mcqa} \label{section:background}

A multiple-choice question (\mcq{}) is a question~$q$ and set of
choices $\mathcal{C}$.\footnote{Some choices can link to many other choices (e.g., ``All of the above''), but these are discouraged~\cite{haladyna2002review}.}
One choice $a \in \mathcal{C}$ is the~gold answer, while others
are plausible-sounding but~incorrect distractors $\mathcal{D} = \mathcal{C}\setminus\{a\}$ meant to test misunderstandings.\footnote{Extractive \qa (e.g., SQuAD) and classification (e.g., \nli{}) can have a finite set of choices, but are fixed (i.e., labels) or have non-misleading distractors, so such tasks are not~\mcqa.}
\mcqa's simple goal---picking the best answer $a$---is popular for \mm{} evaluation, but it has flaws.
Before naming them, we first review its history in
human testing~(\cref{subsection:eval_humans}) and
\abr{nlp}~(\cref{subsection:eval_machines}).

\subsection{Why \mcqa{} is the Standard for Humans}
\label{subsection:eval_humans}

The \mcqa format originated in 1914 with Frederick Kelley's Kansas
Silent Reading Test~\cite{kelly1916kansas}, proposed as an efficient
measure of student reading comprehension~\cite{monroe1917report}.
Soon after, \mcqa was attempted at scale, notably with Robert Yerkes’
Army Alpha and Beta tests~\cite{yerkes1918psychology} in 1917 to assess
U.S. Army intelligence.~An initial bottleneck in \mcqa was the manual effort needed for
scoring, which researchers like Benjamin Wood and Reynold Johnson tackled by designing automated grading systems
~\cite{brennan1971ibm, woodcolumbia} with \abr{ibm}.

Automatic scoring eventually enabled \mcqa's popularity in
primary/secondary education~\cite{butler2018multiple}, college
admissions~\cite{daneman2001using}, language proficiency~\cite{jamieson2000toefl}, and even common tasks like driver's permit exams~\cite{beanland2013there} or employee~training~\cite{puhakainen2010improving}.
%
% \jbgcomment{We should be careful here, as
%   the first College Board tests were essays, and they went to full
%   \mcqa{} later.
%
Parallel to this, education researchers began exploring the best practices for writing high-quality \mcq{}s~\cite{morrison2001writing, campbell2011write}, authoring distractors~\cite{pho2015distractor, gierl2017developing}, and designing test settings~\cite{rakes2008open, shute2013comparison}.

Despite \mcqa's simplicity and popularity, organizations still
critically assess its use in standardized testing.  In the United
States, the \sat{} removes unsound \mcq{}
types,\footnote{https://blog.prepscholar.com/sat-analogies-and-comparisons-why-removed-what-replaced-them} and France's Baccalauréat uses long essay tasks over \mcqa.\footnote{https://www.education.gouv.fr/reussir-au-lycee/le-baccalaureat-general-10457}
We argue \mm{} evaluation needs similar scrutiny and should draw from education to
refine \mcqa's format and data.

% While once popular, \mcqa's dominance in standardized testing is wavering.
% University of California no longer considers \sat{}/ACT scores for admissions,\footnote{https://admission.universityofcalifornia.edu/} and many graduate programs have dropped the GRE,\footnote{https://grenotrequired.com/} citing limited predictive validity.
% As institutions reconsider \mcqa evaluation, we argue similar scrutiny is needed for \mm{} evaluation.

\subsection{How \mcqa Became Popular for \mm{}s} \label{subsection:eval_machines}

\abr{nlp} first used \mcq{}s from human exams; solving these with models that used external sources was considered part of an ``\abr{ai} grand challenge''~\cite{reddy1988foundations}, as it required semantic~\cite{turney2003combining, veale2004wordnet} and factual understanding~\cite{6587172, 10.1145/2509558.2509565}.
Other early \mcq{}s from Winograd~\cite{levesque2012winograd} or COPA~\cite{roemmele2011choice} tested commonsense reasoning over events and ambiguity in premises.
Soon after, \citet{richardson2013mctest} designed MCTest for machine reading comprehension (\mrc{}) via fiction text and \mcq{}s.
All tasks challenged models, but most \mcqa work studied \mrc{}~\cite{lai-etal-2017-race}.

With the advent of larger, neural \lm{}s~\cite{devlin2018bert}, \mcqa needed to become harder.
Researchers expanded \mrc{} to test numerical reasoning~\cite{dua2019drop} and uncertainty~\cite{rogers2020getting}, and successfully scaled existing commonsense \mcq{}s~\cite{sakaguchi2021winogrande}.
New \mcq{}s testing \lm{} pre-training knowledge also grew popular, often using commonsense in daily~tasks~\cite{talmor2018commonsenseqa, bisk2020piqa} and science exams~\cite{mihaylov2018can, clark2018think}.

As \mm{}s improved in generation~\cite{brown2020language}, \mcqa evaluation changed;
models usually scored \mcqa choices independently, but~\citet{robinson2023leveraging} showed prompting \mm{}s with the question and \textit{all} choices was easy to score and matched human testing.
It soon became standard to test \mm{}s with \mcq{}s; companies used~the task to parade their models~\cite{achiam2023gpt}, some equating it to intelligence~\cite{anthropic2024introducing}.
This industry adoption incentivized researchers to write more \mcq{}s across topics~\cite{rein2023gpqa}, languages, and modalities~\cite{zhang2023m3exam}.

% with \mcq{}s being used to test \mm{}s in long-context~\mrc{}~\cite{Pang2021QuALITYQA} and understanding of various subjects~\cite{hendrycks2020measuring}, languages~\cite{hardalov2020exams}, and modalities~\cite{yang2022avqa}.

Recent work critiques \mm{} evaluations generally, discussing reproducibility issues~\cite{laskar-etal-2024-systematic}, how it should be a distinct discipline~\cite{chang2024survey}, and its failure to predict deployment settings~\cite{saxon2024benchmarks}.
We similarly argue that while \mcqa is simple and popular, the task has flaws in its format and datasets, many of which can be fixed using insights from education research.

% \jbgcomment{Last sentence could tie the section together better}

\section{\mcqa is Flawed as a Standard Format} \label{section:format}

\mcqa is a simple format for student testing, but educators find
tradeoffs: it may not predict student success~\cite{moneta2017limitations} or evaluate knowledge~\cite{simkin2005multiple}.
We argue that the same issues apply to \abr{nlp} and thus, \mcqa should not be considered a gold standard for \mm{} evaluation.
Specifically, we discuss \mcqa's rigid goal
(\cref{subsection:best_answer}), misalignment with real \mm{} use
cases (\cref{subsection:use_cases}), and limited testing of knowledge~(\cref{subsection:testing_what}).

\begin{figure}
    \centering
    \includegraphics[width=\linewidth]{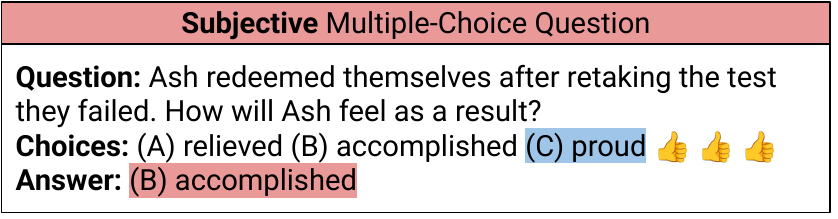}
    \vspace{-4.5ex}
    \setlength{\fboxsep}{0pt}
    \caption{\small Commonsense \mcq from \citet{Palta2024PlausiblyPQ} where \colorbox{bluecolor}{\strut the choice rated most plausible by users} is not the same as the \colorbox{redcolor}{\strut gold answer}; both are subjectively correct in varied contexts.}
    \label{fig:subjective}
    \vspace{-1ex}
\end{figure}

\subsection{``\underline{Pick} the \textit{Best} Answer'' is Too Rigid} \label{subsection:best_answer}

One of \mcqa's key issues is its rigid goal: pick the best answer from a set of choices.
While easy to score, both designs---the use of
1) one gold answer; and 2) input choices---limit \mcqa's applicability.

First, one gold answer hinders \mcqa's use for evaluating subjectivity
\cite{finetti1965methods}.
Still, we use this format for commonsense~\cite{bisk2020piqa}, morals~\cite{yu-etal-2024-cmoraleval, scherrer2023evaluating}, and culture (\cref{subsection:bias}), where many choices can be~subjectively right (Figure~\ref{fig:subjective}).
\citet{Palta2024PlausiblyPQ} find~users rate distractors in commonsense \mcq{}s as the most plausible choice in over 20\% of cases.
Thus, extra care is needed to write \mcq{}s for subjective tasks.

Second, picking from choices means \mcq{}s test \textit{validation}, useful in tasks
like \mm{}-as-a-judge or re-ranking which must compare answers
\cite{gu2024survey}, but inhibiting tasks like writing and coding that
require \textit{generation}~\cite{celikyilmaz2020evaluation}.
One may argue \mcqa proxies generation (if you pick good answers, you generate good ones), but \mm{}s~lack validation/generation consistency
\cite{li2024benchmarking, west2023generative, balepur-etal-2025-reverse}.
Validation and generation are thus separate skills, so \mcqa is a poor format for evaluating generation ability.

In all, \mcqa best tests \mm{}s in objective validation, struggling with subjectivity and generation.

\subsection{Users Rarely Ask \mm{}s to Solve \mcq{}s} \label{subsection:use_cases}

Many leaderboards aim to rank \mm{}s by their overall abilities, helping users select the best model for their needs~\cite{xia2024top}.
Hence, they should adopt tasks that mirror the popularity of user needs, giving higher ranks to models that can actually help users~\cite{balepur-etal-2024-smart, mozannar2025the}.

\mcqa is over-represented versus how \mm{}s are used; 32\% of the tasks in HELM
\cite{perlitz2023efficient}, 71\% in GPT-4's card~\cite{achiam2023gpt}, and 79\% in Open\mm{}~\cite[Big Bench has 21 \mcqa tasks]{open-llm-leaderboard-v2} are \mcqa.
In contrast,~\citet{ouyang2023shifted} find in ShareGPT's set of ChatGPT queries that nearly all user queries
ask for free-form text; we estimate 7.2\% are validation (4.3\% evaluation and 2.9\% comparison).
Similarly, WildChat notes just 6.3\% of their \mm{} queries are factual QA~\cite{zhao2024wildchat}.\footnote{It is not explicitly stated if these are even validation tasks, so this is another \textit{upper} bound of validation task prevalence.}
Thus, over 90\% of queries are likely generative tasks (code, writing, or explanations), which \mcq{}s struggle to test~(\cref{subsection:best_answer}).

% % \jbgcomment{I haven't checked this,
%   but if I remember it correctly, it is predictive, just less
%   predictive than other features.  This is still useful (if I'm
%   remembering correctly) for the story: grad admissions got rid of
%   multiple choice but kept the generative parts.}

Informative evaluation suites must reflect \mm{} use cases.
This is precisely why \mcqa exams are waning in graduate admissions
criteria: they cannot fully predict graduate school
success~\cite{sampson2001gre}. 
Similarly, over-representing \mcqa in evaluations obscures which \mm{}s best aid users.

\subsection{\mcqa Does Not Fully Test Knowledge} \label{subsection:testing_what}

While \mcqa fails to match real user needs (\cref{subsection:use_cases}), we hope the format tests basic skills for such needs, justifying its usage.
\mcqa is meant to test knowledge~\cite{moss2001multiple}, and with input texts, comprehension~\cite{farr1990description}, but work in education shows \mcqa may be suboptimal for these~goals.

\mcq{}s mainly assess the basic knowledge levels in Bloom's Taxonomy of educational goals~\cite{krathwohl2002revision}: recalling, understanding, and applying knowledge~\cite{simkin2005multiple, shin-etal-2024-generation};
it is hard to write \mcq{}s for the higher levels requiring reasoning~\cite{stupans2006multiple, palmer2007assessment, lin2012can}: analyzing, evaluating, and creating knowledge.
As evidence, students can solve \mcq{}s without full understanding, exposed in free-response answers~\cite{mckenna2019multiple}.
\mcq{}s with passages generally test comprehension, but some doubt this;
\citet{ozuru2013comparing} find \mcqa scores correlate with prior knowledge of the passage, overestimating true comprehension.

% In all, \mcqa may reliably test students' comprehension, but not full knowledge assessments.
We believe these same insights can apply to \abr{nlp}: \mcqa may be apt for
comprehension, but rewards \mm{}s for basic recall versus in-depth
knowledge.

\begin{figure}
    \centering
    \includegraphics[width=\linewidth]{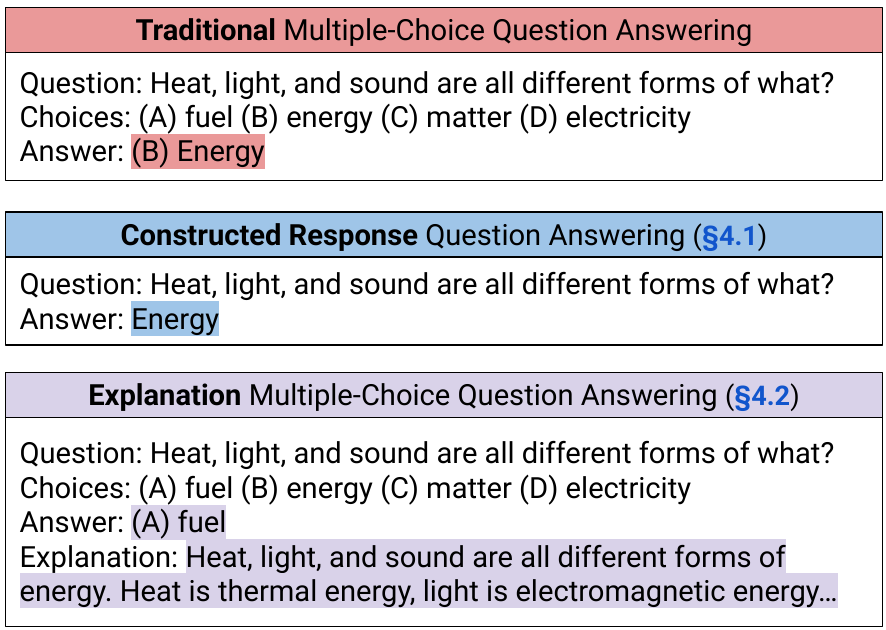}
    \vspace{-4ex}
    \setlength{\fboxsep}{0pt}
    \caption{\small Example of adapting \colorbox{redcolor}{\strut typical MCQs} to our generative formats: \colorbox{bluecolor}{\strut Constructed Response} and \colorbox{purplecolor}{\strut Justified \mcqa}.}
    \label{fig:format}
    \vspace{-1ex}
\end{figure}

% Normal texts are the task inputs, while highlighted texts are GPT-4o's responses.

\section{\underline{Generative} \mcqa Tasks are Promising} \label{section:fixing_format}

\mcqa is flawed, so how should its role change?
Standardized LLM evaluations must 1) proxy LLM use cases; and 2) test skills for (1).
\mcqa is unfit for (1), so~evaluations need more tasks matching LLM needs (\cref{subsection:use_cases}).
Writing/explanation tasks are harder to score \cite{charney1984validity, chakrabarty-etal-2022-help, balepur-etal-2023-expository}, but it is still odd they are often omitted, as they are typical needs~(\cref{subsection:use_cases}).

This limits \mcqa to (2), but it is best for comprehension or validating objective facts (\cref{subsection:best_answer}), not generation, subjectivity, or knowledge (\cref{subsection:testing_what}).
Validation/comprehension are valuable, which is why we discuss \mcqa datasets in \cref{section:dataset}, but we need~better formats for the other skills.
Thus, we give two generative versions of \mcqa to better test LLMs (Figure~\ref{fig:format}):
Constructed Response (\cref{subsection:constructed_response}), answering sans choices, and Explanation \mcqa (\cref{subsection:justified_mcqa}), justifying predictions.
We ensure the formats only slightly increase evaluation complexity, mostly preserving \mcqa's simplicity of scoring. 
We now~describe the formats and future work to realize them.

\subsection{Constructed Response Questions} \label{subsection:constructed_response}

Having LLMs solve MCQs without choices, called \textbf{Constructed Response (CR)} in education \cite{livingston2009constructed} or short-form \qa in \abr{nlp} \cite{krishna-etal-2021-hurdles}, is one better format.
CRQs test answer \textit{generation} unlike MCQs (\cref{subsection:best_answer}), better mirroring LLM needs (\cref{subsection:use_cases}), and it is easier to write CRQs testing all skills in Bloom's Taxonomy \cite{krathwohl2002revision}, so they better expose knowledge gaps (\cref{subsection:testing_what}).
Thus, students find CRQs harder than MCQs~\cite{hancock1994cognitive}, which can also delay saturation (\cref{subsection:saturation}).

Instead of writing CRQs to replace our vast existing \mcqa data, a promising solution is to convert MCQs into CRQs by omitting choices and tasking LLMs to give a \textbf{short-form} answer $\hat{a}$ for question $q$, comparing it to gold answer $a \in \mathcal{C}$ \cite{bhakthavatsalam2021think}.
\citet{myrzakhan2024open}~show two hurdles in this: finding MCQs to convert\footnote{\;``Which of these best...'' MCQs require using all choices.} and scoring $\hat{a}$ with $a$. 
Recent efforts in flagging MCQ errors \cite{moore2023assessing, moore2024automatic} and judging short-form answer correctness \cite{li-etal-2024-pedants, moore2022assessing} showcase that we can realistically overcome these challenges.
Thus, we~believe~combining this work with best practices for creating~CRQs \cite{snow2012construct} can successfully implement the task.

% Reliable MCQ to CRQ conversion would thus require combining advancements in MCQ generation \cite{lee2024math} and automatic scoring \cite{li-etal-2024-pedants} with best practices for writing and evaluating CRQs~\cite{snow2012construct}.

\subsection{Explanation Multiple Choice Questions} \label{subsection:justified_mcqa}

Constructed Response is promising (\cref{subsection:constructed_response}), but using one short answer is unfit for subjectivity \cite{lin2020differentiable} and conflicts user preferences for long outputs \cite{zheng2024judging}.
We thus propose~\textbf{Explanation \mcqa (\emcqa)} as another \mcqa alternative from education \cite{lau2011guessing}: for a question $q$ and choices $\mathcal{C}$, models give an answer $\hat{a} \in \mathcal{C}$ and explanation $\mathcal{E}$ for why $\hat{a}$ is right.
This format tests generation (\cref{subsection:best_answer}), matches the use case of explanations (\cref{subsection:use_cases}), and has shown to test more knowledge levels over \mcqa \cite{lee2011validating}.

%This format is also supported by education research, as asking for justifications can expose student knowledge gaps and heighten task difficulty \cite{tamir1990justifying, koretsky2016written}.

We envision \emcqa being treated like reasoning tasks \cite{cobbe2021training}, checking if LLMs pick $a$ like \mcqa's simple scoring, but also studying $\mathcal{E}$.
If models select $a$ but justify it poorly, it exposes knowledge gaps like in student assessments \cite{jonassen2010arguing}, and when models give strong explanations for wrong answers, it enables partial credit for subjective tasks \cite{lau2011guessing}.

\emcqa has many benefits, but needs metrics to score ``good'' explanations over many facets \cite{xu2023critical}, like factuality to curb hallucinations \cite{min2023factscore, balepur-etal-2023-text}, plausibility for convincingness \cite{liu2023vera}, and faithfulness to verify $\mathcal{E}$ supports $a$ \cite{lanham2023measuring}.
We believe these goals could be achieved by merging ongoing efforts in building verifiers for LLM reasoning \cite{NEURIPS2023_72393bd4} with educational best practices for grading justifications \cite{jonassen2010arguing}, yielding reliable metrics that realize \emcqa's potential.
\citet{kim-etal-2025-biggen} have made notable progress for this goal---building LLM judges to score justifications across various benchmarks---showing that implementing E-MCQA is feasible.

\section{\mcqa Datasets are Flawed but Fixable} \label{section:dataset}

\mcqa is not always the best format (\cref{section:format}), but~we still need high-quality \mcq{}s for comprehension/validation as well as tasks like \mm{}-as-a-judge~\cite{gu2024olmes} and re-ranking~\cite{ma-etal-2023-large} which require comparing answers.
Further, our generative \mcqa formats (\cref{section:fixing_format}) still use \mcq{}s as inputs.

However, like most \abr{nlp} tasks, \mcqa datasets have quality issues that impede their utility: leakage
(\cref{subsection:test_set_leakage}), unanswerability
(\cref{subsection:quality}), shortcuts (\cref{subsection:artifacts}),
and saturation (\cref{subsection:saturation}).
We now show how educators' solutions to these issues can inform \abr{nlp} datasets.

% \jbgcomment{I think it's better for filenames to be named so that they sort correctly}

\subsection{\mm{}s Peek at \mcqa Answer Keys} \label{subsection:test_set_leakage}

To build an \mcqa dataset, we first need sources to write or collect
\mcq{}s.
But as many sources end up being leaked\footnote{gpt-3 has seen
45\% of RACE's test set~\cite{sainz2023nlp}.} in \mm{} training data
\cite{magar2022data}, such \mcq{}s may confuse~generalization abilities for
memorization~\cite{lewis2020question}.
Private test sets~\cite{sap2019socialiqa} and de-contamination
\cite{zhou2023don} help, but \mm{}s tuned on newer data can overlap with
(1), and opacity in \mm{} data~\cite{soldaini2024dolma} blocks (2).
% \jbgcomment{Patrick Lewis's paper on NQ overlap is probably worth citing here}

An ambitious solution to test set leakage is live \mcq{}s that update over time to stay unseen~\cite{white2024livebench}, like how educators rewrite exams to impede cheating. Trivia~\cite{jennings2007brainiac}~and standardized testing groups frequently write new questions, making them ideal partners.
To aid both~parties, researchers could offer these groups tools for tutoring~\cite{li-etal-2024-eden}, \mcq{} validation~\cite{yu-etal-2024-automating-true}, or answer scoring~\cite{yang-etal-2020-enhancing}.

Test set leakage would be easier to fix if model designers released
training data, but as we all know, most do not.
While it is harder to design solutions for leakage that do not need training data, we hope researchers view it as
a challenging, impactful research problem in evaluation and
generalization.

\subsection{Some \mcq{}s Have No Correct Answer} \label{subsection:quality}

Once a source is found (\cref{subsection:test_set_leakage}), researchers collect~or write \mcq{}s, but errors often arise rendering them unanswerable, like mislabeling~\cite{explained2023smart}, multiple correct choices~\cite{Palta2024PlausiblyPQ}, ambiguity~\cite{gema2024we}, missing contexts~\cite{wang2024mmlu}, and grammar errors~\cite{chen2023hellaswag}.

\begin{figure}
    \centering
    \includegraphics[width=\linewidth]{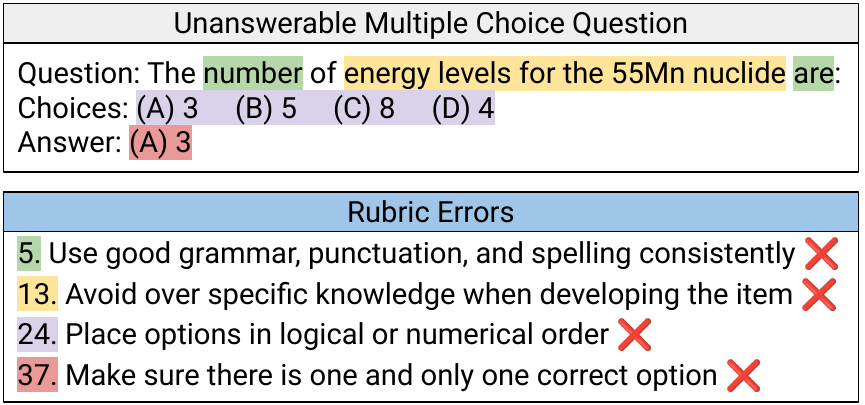}
    \vspace{-4ex}
    \setlength{\fboxsep}{0pt}
    \caption{\small Example unanswerable MCQ from MMLU \cite{gema2024we}, along with rubric criteria from \citet{haladyna1989taxonomy} flagged by OpenAI's o1 \cite{jaech2024openai}.}
    \label{fig:checklist}
    \vspace{-1ex}
\end{figure}

Educators write \mcq{}s with rigorous protocols, and we must meet similar standards in \abr{nlp}~\cite{boyd2019question}; we should use educators' rubrics (Figure~\ref{fig:checklist}) for writing and validating \mcq{}s~\cite{haladyna1989taxonomy}.
Such guidelines also specifically exist for distractors~\cite{haladyna2002review}---the part of \mcq{}s that discern testees' skills---ensuring they are truly wrong, shortcut-proof (\cref{subsection:artifacts}), and not too easy to rule out (\cref{subsection:saturation}).
Beyond \mcq{} writing, rubrics can form~data cards~\cite{pushkarna2022data} to help researchers record errors in their data and how~they~fixed them.

Recent work in \mm{} checklist evaluation~\cite{cook2024ticking}, 
\mcq{} metrics~\cite{moon2022evaluating},~and \mcq{} generation~\cite{feng-etal-2024-exploring} show parts~of this workflow can be
automated (Figure~\ref{fig:checklist}).
\citet{wang2024mmlu} fix errors in \mmlu by using \mm{}s to detect issues and
write new choices.~\mm{} judges can be inaccurate and biased~\cite{xu-etal-2024-pride}, so
human-\abr{ai} collaboration, like model-assisted
refinement~\cite{shankar2024validates} and task
routing~\cite{miranda2024hybrid}, may be more promising.
Errors will arise in \mcq{} writing, but educators' rubrics can help
find and fix them, ensuring answerability.

\subsection{\mcqa Shortcuts May Let \mm{}s ``Cheat''} \label{subsection:artifacts}

Answerable \mcq{}s (\cref{subsection:quality}) are not always high quality, as shortcuts may let \mm{}s guess the answers to \mcq{}s without knowing the answers~\cite{10.1145/3596490}, overestimating model accuracy~\cite{wiegreffe2021teach}.
Shortcuts arise from annotator artifacts~\cite{gururangan2018annotation}, spurious patterns~\cite{zhou2023explore}, or bypassed reasoning steps~\cite{chen2019understanding}.
If \mm{}s best random guessing via partial inputs~\cite[e.g., choices only]{richardson2020does}, they exist.

Below, we discuss how scoring (\cref{subsubsection:shortcut_scoring})
and data collection methods (\cref{subsubsection:shortcut_dataset})
can mitigate shortcuts.

% \jbgcomment{Note for the future: if YooYeon's AdvCal paper gets in, that should be cited here (or in RW) as well}

\subsubsection{Calibrated Scoring Can Deter Guessing} \label{subsubsection:shortcut_scoring}

While rare in human testing,\footnote{Since
they may induce stress or anxiety in students~\cite{vanderoost2018elimination},
but we think it is fine to stress~\mm{}s.} scoring methods can
penalize wrong guesses~\cite{lau2011guessing}:
\textbf{1) Probability scoring:} elicit confidence scores for each choice~\cite{finetti1965methods}; \textbf{2) Negative marking:} subtract points for wrong answers with abstention allowed~\cite{holt2006analysis}; and \textbf{3) Elimination scoring:} students iteratively remove wrong choices until unsure~\cite{ben1997comparative}.
Confidence~\cite{li-etal-2024-think}, abstention~\cite{goral2024wait}, and elimination~\cite{ma2023poe} have been studied in \mm{}s, so they may be easy to use for \mcqa evaluation.

Calibration methods deter guessing, but also reward models that know their knowledge gaps~\cite{guo2017calibration}.
Such scoring methods are often ignored in evaluation~\cite{bommasani2023holistic}, but it could let \mcqa better test decision-making~\cite{liu2024dellma} and enable partial credit for subjective tasks where many choices may be tenable~(\cref{subsection:testing_what}).

\subsubsection{We Can Design Shortcut-Proof \mcq{}s} \label{subsubsection:shortcut_dataset}

Data designers should limit shortcuts; an easy way is
uniform design.
When solving \mcq{}s with only choices,
\citet{Balepur2024ArtifactsOA} find \mm{}s may exploit distributional
differences, so like educators~do (\cref{subsection:quality}), we should write parts of \mcq{}s consistently: via the same agent, source, and decoding method.
HellaSwag~\cite{zellers2019hellaswag} leads to the highest known choices-only accuracy, where \textit{user}-written answers and \textit{model}-written distractors are inconsistent, showing the necessity of uniform data design.

\begin{figure}
    \centering
    \includegraphics[width=\linewidth]{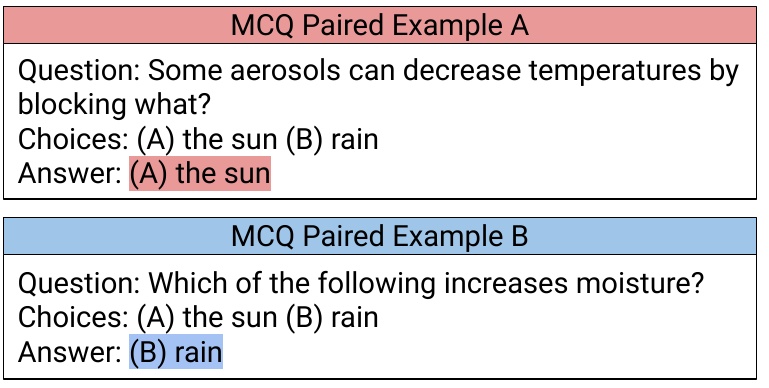}
    \vspace{-4ex}
    \setlength{\fboxsep}{0pt}
    \caption{\small Example MCQ pair for a contrast set from \citet{balepur2024your}. The choices are identical, but the question swaps the answer, testing if models ignore the question.}
    \label{fig:contrast}
    \vspace{-1ex}
\end{figure}

Contrast sets are another tool that detects if models ignore inputs and use shortcuts~\cite{Gardner2020EvaluatingML}.
In \mcqa, they are entry pairs differing by some inputs (e.g., question)
that change the answer (Figure~\ref{fig:contrast}), ensuring models attend to the perturbed input
\cite{elazar2023measuring}.
\citet{balepur2024your} use contrast sets in commonsense \mcqa to
ensure none of their \mm{}s rely on shortcuts in choices to rank
highly.
Contrast sets are often made manually~\cite{Kaushik2020Learning}, so future work~can test
automatic ways to build them~\cite{li-etal-2020-linguistically}.

% \jbgcomment{self cites are a little prevalent here, pare back if you can}

%~\cite{srikanth-rudinger-2022-partial}

Studying how users and models \textit{``cheat''}
\cite{saxon-etal-2023-peco} in \mcq{}s also finds shortcuts.
In reading comprehension,~\citet{Pang2021QuALITYQA} give~users \inlinecode{ctrl+F} to detect \mcq{}s quickly solvable without using the full text, while~\citet{Malaviya2022CascadingBI} flag \mcq{}s where users
can use simple heuristics.
We can also train models to cheat; adversarial
filtering trains simple models
(e.g., bag-of-words) and omits \mcq{}s they can solve~\cite{zellers2018swag}.~Extending this, we believe having strong \mm{}s reason to cheat---similar to safety work in alignment faking
\cite{greenblatt2024alignment}---can help find shortcuts.

\subsection{\mm{}s Inevitably Ace \mcqa Datasets} \label{subsection:saturation}

Even if we fix all of these issues, \mcq{}s become too easy over
time (i.e., saturated), no longer tracking \mm{} progress
\cite{li2024crowdsourced}.
To still use ``easy'' \mcqa datasets, we need to
make them harder for models.
%
% \jbgcomment{this sentence could be more direct: ``propose'' and ``study'' are filler words in academic writing}
Below, we show how understanding which \mcq{}s
are hard (\cref{subsection:irt}) and helping users author hard, interpretable \mcq{}s (\cref{subsection:hitl}) delay saturation.

\subsubsection{\irt{} Reveals Challenging \mcq{}s} \label{subsection:irt}

When \mm{}s excel in \mcqa datasets, some \mcq{}s remain hard; finding
them and why they are hard informs data design
\cite{sugawara2018makes, Sugawara2022WhatMR}.
One \mcq{} difficulty metric is success rate (\success)---the number of models
answering correctly~\cite{gupta2024improving}---but \success omits
\textit{which} models succeed.
\mcq{}s solved by just the worst or best model have equally low success rates, but the former suggests \mcq{} errors (\cref{subsection:quality})---as a weaker model besting all others is rare---while the latter matches our expectation. 
As \success conflates these cases, it cannot separate flawed \mcq{}s from those discerning model~ability.

% \jbgcomment{I think the better argument that \irt{} is better than SR is
%   that bad questions can have a low success rate: you want to
%   eliminate questions with errors, ambiguities, etc.  And leave the
%   discriminative questions.  You already introduced why you want
%   discriminative questions, so you can just back point to that.  I'd
%   talk more about the failure modes of SR, which can highlight how the
%   opposite of those are things \irt{} can find.}

Item Response Theory (\irt{})---a tool used in education~\cite{lord2008statistical}---is a more robust way to find hard \mcq{}s.
While \success treats all~models equally, \irt{} learns the skill of models to then estimate every \mcq{}'s difficulty (how hard it is) and discriminability (how well it discerns between weak/strong models).
\irt{} can then be used to~filter high-difficulty, high-discriminability \mcq{}s as harder data splits~\cite{polo2024tinybenchmarks}, flag saturation if all \mcq{}s have low difficulty/discriminability~\cite{vania2021comparing}, and omit faulty \mcq{}s with negative discriminability~\cite{rodriguez-etal-2021-evaluation}.

% rodriguez-etal-2021-evaluation

While \irt{}-based filtering finds harder \mcq{}s, it does not give new \mcq{}s, limiting its long-term use.
However, we can extend \irt{} to multi-dimensional \irt{}~\cite[M\irt{}]{reckase200618} to capture \textit{many} latent skills, offering more insights into model abilities; by interpreting these skill dimensions, we can pinpoint model issues that inform future data efforts (\cref{subsection:hitl}).
\citet{gor2024great} reveal \mm{} errors in abductive reasoning via a variant of M\irt{}---an issue confirmed by abduction research~\cite{del-fishel-2023-true, DBLP:conf/iclp/NguyenGTSS23}.
M\irt{} could similarly find difficult \mcqa topics, distractor patterns, or reasoning types for models~\cite{benedetto-etal-2021-application}.

%~\cite{balepur2024reverse}

% \jbgcomment{Density of self-cites a little high here (for the group, not just you)}

Overall, \irt{} can find which \mcq{}s are hard but also why, informing future data collection efforts.

% \jbgcomment{Section title doesn't cover adversarial questions that well.  Perhaps a transition of the form: while engaged authors are used to understanding what trips up humans, they may need interface / computational support to trip up models\dots

% That could transition to adversarial data.}

\begin{figure}
    \centering
    \includegraphics[width=\linewidth]{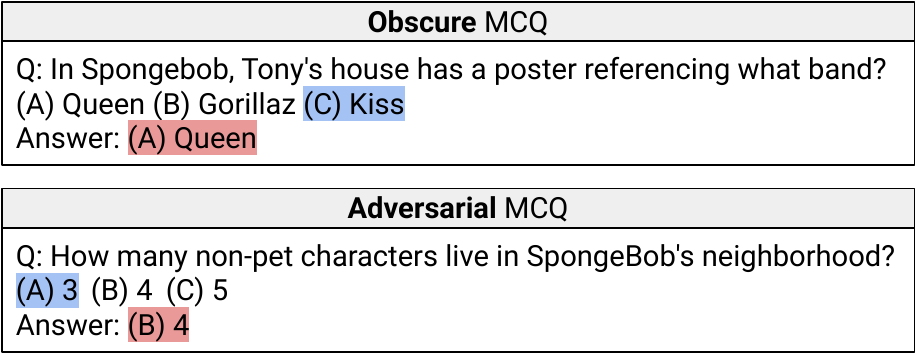}
    \vspace{-4ex}
    \setlength{\fboxsep}{0pt}
    \caption{\small Obscure and adversarial MCQs for \textit{Spongebob Squarepants} inspired by \citet{sung2024your}. GPT-4o answers both \colorbox{redcolor}{\strut wrong} (answer in \colorbox{bluecolor}{\strut blue}). The former tests niche knowledge, but the latter is easy for those who have seen the~show.}
    \label{fig:adv}
    \vspace{-1ex}
\end{figure}

\subsubsection{A Good \mcq{} is Hard \textit{and} Interpretable} \label{subsection:hitl}

A popular way to write harder \mcq{}s is requiring obscure knowledge
(Figure~\ref{fig:adv}, left), sourcing from experts
\cite{rein2023gpqa, phan2025hle} and global competitions
\cite{fang2024mathodyssey}.
These challenge models \textit{and} humans, which is useful for \abr{ai} safety work in scalable oversight~\cite{bowman2022measuring}.
However, this makes them uninterpretable for non-experts diagnosing model errors, especially when studying model rationales (\cref{subsection:explanations}).
If \mm{}s~err on obscure \mcq{}s, it is hard for non-experts to find if errors are from faulty reasoning, misunderstandings, or knowledge gaps~\cite{anderson2025phdknowledgerequiredreasoning}.

While authors know which \mcq{}s elude humans, writing ones that surface model errors while staying human-interpretable needs support.
This is \textbf{adversarial} data collection's goal~\cite{kiela-etal-2021-dynabench}---building \ui{}s to help authors write examples hard for models but easy for humans (Figure~\ref{fig:adv}, right).
Rather than using niche knowledge, authors must write \mcq{}s with spurious patterns, misleading distractors (\cref{subsection:quality}), or reasoning traps that trick \mm{}s but not humans~\cite{xu2023llm}.
As a result, these \mcq{}s better expose robustness (\cref{subsection:robustness}) and logical reasoning (\cref{subsection:explanations}) errors, less clouded by knowledge.

Adversarial \mcq{}s are useful, but finding users to write high-quality ones is tough.
Gamification---making the task fun---helps, used in building adversarial commonsense~\cite{talmor2022commonsenseqa}, \qa~\cite{wallace2019trick, sung2024your}, and fact-checking \cite{wallace2019trick} data.
\citet{wallace-etal-2022-analyzing} use gamification to get adversarial \nli{} data and show it has \textit{long-term} difficulty, delaying saturation. 
%~\citet{sung2024your} recently use gamification with \irt{} (\cref{subsection:irt}) to build an adversarial QA dataset with the highest gap in human and model accuracy in the past five years.
For more engagement, researchers can stir users to author \mcq{}s exposing shocking failures, inspired by jailbreaking, where provocative outputs are naturally fun to elicit~\cite{schulhoff2023ignore}.
%\cite{schulhoff2023ignore}.

% \jbgcomment{DADC would be better for submission (the barrage of self-cites can return for camera ready, so don't delete, jsut comment)}

Obscure and adversarial \mcq{}s can both make \mcqa harder: the former
tests niche knowledge, while the latter better find reasoning or
consistency failures that are uneclipsed by knowledge gaps.

\begin{figure}
    \centering
    \includegraphics[width=\linewidth]{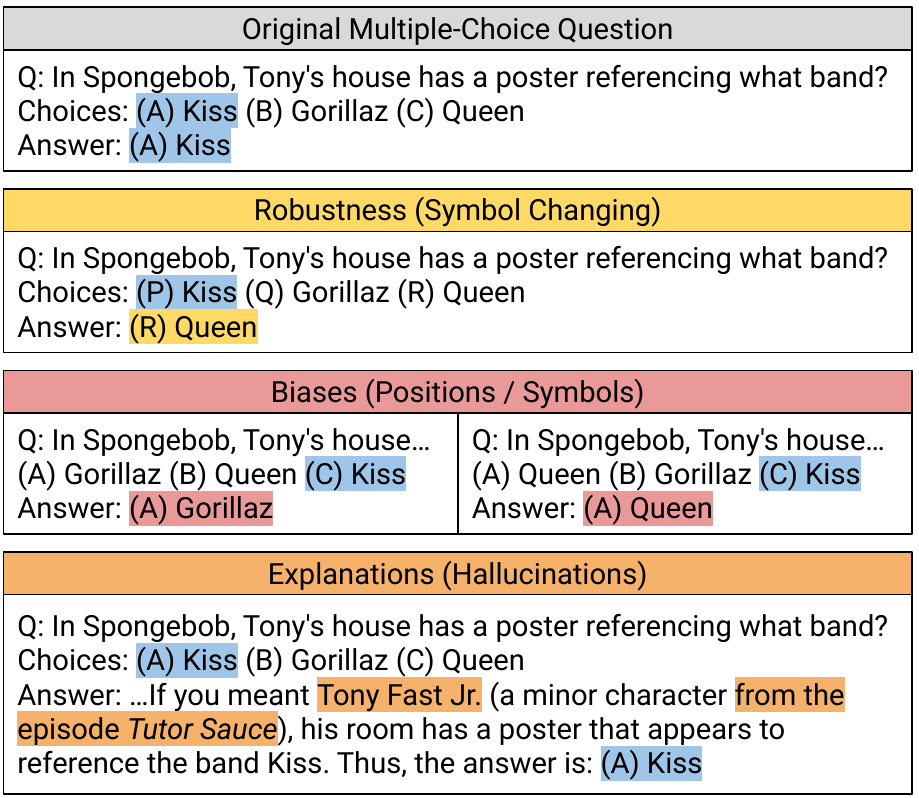}
    \vspace{-4ex}
    \setlength{\fboxsep}{0pt}
    \caption{\small LLMs err with \colorbox{yellowcolor}{\strut robustness} (e.g. inconsistent after shuffling), \colorbox{redcolor}{\strut biases} (e.g. favor symbols), and \colorbox{orangecolor}{\strut explanations} (e.g. fail to justify answers) in \mcqa. Correct answers are \colorbox{bluecolor}{\strut blue}.}
    \label{fig:llm_issue}
    \vspace{-1ex}
\end{figure}

\section{Fixing \mcqa Can Help Us Fix \mm{}s} \label{section:models}

Fixing issues in \mcqa's format (\cref{section:format}) and datasets (\cref{section:dataset}) will not just improve evaluation quality; they can also improve our understanding of \mm{} weaknesses.
This section outlines three persistent
issues of \mm{}s in \mcqa (Figure~\ref{fig:llm_issue})---robustness~(\cref{subsection:robustness}), bias
(\cref{subsection:bias}), and explanations
(\cref{subsection:explanations})---and how our prior solutions can better address or evaluate them.

\subsection{New Prompts Lead to New \mcqa Scores} \label{subsection:robustness}
% \subsection{\mm{}s Lack Robustness in \mcqa} \label{subsection:robustness}

% On specific prompts, \mm{}s score highly in~\mcqa, but currently crumble
% when prompts change, sensitive to: choice
% symbols~\cite{alzahrani2024benchmarks}, choice
% order~\cite{pezeshkpour2023large, Zheng2023LargeLM}, and
% phrasing~\cite{holtzman-etal-2021-surface, Wiegreffe2023IncreasingPM}.
% This brittleness weakens \mcqa leaderboard reproducibility, as varied
% setups yield conflicting rankings~\cite{gu2024olmes}

On specific prompts, \mm{}s score highly in~\mcqa, but now crumble
after prompts change, sensitive to: choice
symbols~\cite{alzahrani2024benchmarks}, choice
ordering~\cite{Zheng2023LargeLM}, and
phrasing~\cite{Wiegreffe2023IncreasingPM}.
This brittleness degrades \mcqa leaderboard reproducibility, as different evaluation setups yield conflicting rankings~\cite{gu2024olmes}

\mm{} robustness varies by task setup.
In \mcqa, early \mm{}s with poor instruction
following~\cite{zhang2022opt} used probability-based \textit{scoring},
while instruction-tuning enabled \mm{}s to \textit{generate}
answers~\cite{longpre2023flan}; while logically equivalent, these give varying answers~\cite{lyu2024beyond}. 
Probability scoring seems more at fault---more sensitive to prompts~\cite{wang2024look}---doubting if \mm{}s can aid decision-making tasks that need accurate confidence scores~\cite{liu2024dellma}.
To track this progress in \mm{}s, researchers can use \mcqa scoring protocols that measure calibration~(\cref{subsubsection:shortcut_scoring}).

As accuracy often \textit{drops} post-perturbation~\cite{zhou-etal-2024-revisiting}, these errors in generalization could indicate dataset leakage (\cref{subsection:test_set_leakage}) or over-reliance on biases (discussed in \cref{subsection:bias}).
Another proposed explanation is symbol binding error: \mm{}s ``know'' the answer but cannot link it to the right choice~\cite{Wiegreffe2024AnswerAA, xue2024strengthened}.
These failures weaken \mcqa's ability to evaluate knowledge, obscured by memorization and symbol binding.\footnote{Such evaluations have downstream use (e.g., privacy, interpretability), but do not measure knowledge as intended.}~Our proposed solutions---like curating live \mcq{}s for data leakage (\cref{subsection:test_set_leakage}) and generative \mcqa formats to expose knowledge gaps without needing symbol binding~(\cref{section:fixing_format})---can more reliably assess knowledge.

% \jbgcomment{If short on space, some of these could be move to appendix}

\subsection{\mm{}s are Biased \mcqa Test-Takers} \label{subsection:bias}

Like most \abr{nlp} tasks~\cite{baumler2022recognition, chu2024fairness}, \mm{}s show biases in \mcqa, grouped into two main types.
%
% \jbgcomment{can we be a little more precise than ``favor''?}
%
The first are \mcqa-specific biases---selecting answers based on symbols~\cite{Zheng2023LargeLM}, positions~\cite{Li2024CanMQ,
wei-etal-2024-unveiling}, or phrases such as ``none of the
above''~\cite{xu2024llms, wang2025llms} rather than \mcq{} content.
This degrades robustness
(\cref{subsection:robustness}), masking model knowledge.
We believe they likely stem from shortcuts (\cref{subsection:artifacts});
\mm{}s tuned on data where choices with patterns are often
correct will manifest these biases~\cite{pacchiardi2024leaving}.
Our safeguards---uniform design, contrast sets,
and ``cheating'' (\cref{subsubsection:shortcut_dataset})---can
reduce these biases, and more work in shortcuts may be key
to quash them.

The second bias type stems from general training, like cultural/linguistic bias~\cite{myung2024blend, li2023land};
\mm{}s often err on non-Western cultural \mcq{}s~\cite{Acquaye2024SusuBO, azime2024proverbeval} and non-English \mcq{}s~\cite{son2024kmmlu, li2023cmmlu}.
\mcq{}s are popular for assessing bias~\cite{guo2023evaluating}, but if they use subjective commonsense~\cite{seo-etal-2024-kocommongen}, they are subpar (\cref{subsection:best_answer}).
Thus, we advise writing \mcq{}s with rubrics to limit ambiguity and correct distractors (\cref{subsection:quality}),~ensuring bias is more objectively tested.
If subjectivity persists, evaluators should consider strategies to manage it like Explanation \mcqa~(\cref{subsection:justified_mcqa}) or calibration scoring (\cref{subsubsection:shortcut_scoring}), aiding bias testing and for \emcqa, better matching how bias presents downstream~\cite{seshadri2025does}.

Lastly, to construct non-English \mcq{}s, we discover researchers typically either: 1) collect \mcq{}s in the desired target language~\cite{son2024kmmlu}; or 2) translate English ones~\cite{achiam2023gpt};
However, (1) may be infeasible for languages that are low-resource or where \mcqa is rarely used,\footnote{In Germany it is uncommon to test students with \mcq{}s: https://www.reddit.com/r/AskAnAmerican/comments/rjxns4/} while (2) may introduce error propagation during translation~\cite{singh2024global}---showing neither method is perfect on its own.
Thus, researchers can adopt our prior solutions for both methods---like using rubrics during translation (\cref{subsection:quality}) or collecting non-English \mcq{}s from unseen sources (\cref{subsection:test_set_leakage})---improving evaluations of multilingual abilities.

\subsection{\mm{}s Struggle to ``Explain Their Work''} \label{subsection:explanations}

% \mm{}s can easily pick \mcq{} answers~(\cref{subsection:saturation}), but often give \textbf{unfaithful} explanations---failing to mirror their true reasoning---whether via chain-of-thought~\cite{lyu2023faithful, lanham2023measuring, turpin2024language} or self-explanations~\cite{agarwal2024faithfulness, kim2024can, madsen2024self}.
% However, they are convincing, surpassing crowdworkers in perceived quality~\cite{mishra-etal-2024-characterizing} and misleading humans when incorrect~\cite{si2023large}, likely due to \mm{} alignment protocols that optimize user preferences instead of accuracy~\cite{wen2024language}.

Even if \mm{}s get the right answer to an~\mcq{}, they may justify their selection \textbf{unfaithfully}---failing to mirror their true reasoning---whether via chain-of-thought~\cite{lyu2023faithful, turpin2024language} or self-explanations~\cite{kim2024can, madsen2024self}.
However, they are convincing, besting crowdworkers in judged quality~\cite{mishra-etal-2024-characterizing} and misleading users when wrong~\cite{si2023large}, likely as standard alignment strategies tend to optimize \mm{}s for user preferences over helpfulness~\cite{balepur-etal-2024-smart, wen2024language}.

\mm{} explanation flaws are even clearer after logical consistency checks.
\citet{kawabata2023evaluating} show \mm{}s often inaccurately explain answers to subquestions in reading comprehension, even when answering the higher-level question correctly.
Similarly,~\citet{balepur-etal-2024-easy} find \mm{}s struggle to reason why distractors are wrong.
These issues likely stem from broader \mm{} logical inconsistencies~\cite{liu2024aligning, varshney2024investigating}.

Explanations are popular \mm{} use cases~(\cref{subsection:use_cases}), highlighting the need for improved \mcqa formats such as Explanation \mcqa (\cref{subsection:justified_mcqa}) which explicitly test explanation skills.
Further, \mm{}s' logically inconsistent explanations offer a path toward harder datasets (\cref{subsection:saturation}); tools like MIRT (\cref{subsection:irt}) can identify logical error types (negation, decomposition) that elude \mm{}s, while adversarial collection can curate \mcq{}s to excise these errors while staying easy for humans (\cref{subsection:hitl}).
In all, \mm{}s' poor explanation abilities give an opportunity to design harder \mcqa evaluations better aligned with user needs.

% \subsection{Shortcuts}

% Word level shortcuts in \mcqa~\cite{pacchiardi2024leaving}

% BERT can do MRC without the passage, random keywords distract BERT, perturbations change accuracy~\cite{si2019does}

% More MRC shortcuts~\cite{Yu2020CounterfactualVC}

% Quantifying the learnability of shortcuts~\cite{Shinoda2022WhichSS}

% MRC doesnt need the input passage~\cite{raina2023analyzing}

% MRC doesnt need the input passage, neither do humans~\cite{liusie2022world} 

% Easy to cheat on \mcqa versions of multi-hop datasets~\cite{chen2019understanding}

% \subsection{Future Directions}

% Interpretability via contrastive edits~\cite{ross2020explaining}

% Interp via circuits~\cite{lieberum2023does}

% Interpreting biases~\cite{shen2022understanding}

% AI Safety --- some \mcqa tasks are hard for humans, we can study models in relation to helping humans~\cite{parrish2022single, parrish2022two, wen2024language}

% \section{Related Work}

% Survey on MRC \cite{foolad2024recent}

% Survey on LLM cognitive biases \cite{sumita2024cognitive} (like order bias)

\section{Call to Action: Benchmarking 101} \label{section:conclusion}

% \jbgcomment{I think this ``call to action'' could be a little stronger and more specific.}

If you want to make the best benchmark ever, where do you begin?
First, define the ability you want to test and decide if \mcqa is the right format~(\cref{section:format}).
If the ability matches a downstream task (e.g., coding), just use that task~\cite{saxon2024benchmarks}.
If~the ability is fundamental (e.g., knowledge), consult education research to weigh alternative formats~(\cref{section:fixing_format}).

If \mcqa is the best format, find a data source~to curb leakage---one with fresh content (\cref{subsection:test_set_leakage}).
When curating \mcq{}s from your source,~follow educators' rubrics to ensure answerability (\cref{subsection:quality}), and release the rubric as a data card to record errors~\cite{pushkarna2022data}.
Consistent design choices will limit shortcuts (\cref{subsection:artifacts}), and providing a contrast set could help researchers check if their models over-rely on shortcuts~\cite{Gardner2020EvaluatingML}.
As another safeguard, your benchmark can use calibration scoring beyond accuracy to discourage guessing~(\cref{subsubsection:shortcut_scoring}).

Post-release, models will hill-climb and saturate your data over time (\cref{subsection:saturation}).
If you want to delay~saturation, you may restart with an obscure knowledge source, but if you want your data to better diagnose errors, aim for interpretability.
Use IRT (\cref{subsection:irt}) to find which of your \mcq{}s are hard and why, then design an engaging, adversarial dataset collection protocol (\cref{subsection:hitl}) guided by these insights, yielding a new dataset hard for models but easy for humans.

By using even some of educators' insights, we can refine the utility of \mcqa---or any task.
This approach takes more effort than the simple \mcqa practices that initially attracted researchers, but if we do not address the flaws of \mcqa, \textbf{what~model abilities can our \mcqa benchmarks even test?}

\section*{Acknowledgments}

We wish to thank the \abr{clip} lab at the University of Maryland and external collaborators for their help.
In particular, we thank Paiheng Xu, Dang Nguyen, Shi Feng, Ioana Baldini, Vipul Gupta, and the Google Translate Reading Group for general discussions; Yu Hou, Dayeon Ki, and Connor Baumler for discussions of culture/biases; Maharshi Gor for feedback on \irt; Yoo Yeon Sung for feedback on \irt and adversarial data collection; Matthew Shu for feedback on best practices in educational testing; Atrey Desai for helping us test MCQ checklist evaluation; and Naina Balepur for brainstorming our excellent title.
We sincerely appreciate Michael Saxon, Yi Ting Huang, Fumeng Yang, and Andrew Lan for their reviews of earlier versions of this paper.
This material is based upon work supported by the National Science Foundation under Grant No. \abr{iis}-2403436 (Boyd-Graber), \abr{iis}-2339746 (Rudinger), and \abr{dge}-2236417 (Balepur).
Any opinions, findings, and conclusions or recommendations expressed in this material are those of the author(s) and do not necessarily reflect the views of the National Science Foundation.

\section{Limitations} \label{section:limitations}

Inspired by~\citet{saxon2024benchmarks}, we organize our limitations section as potential counterarguments:

\paragraph{I Do Not Work on \mcqa:}

Our approach to \mcqa can apply to all tasks;
it is important to question if your format effectively evaluates your intended ability.
Educators have long studied the best formats for different tasks, but we are not using these insights to guide our benchmark design.
As an example beyond \mcqa, in human math assessments, students are often required to ``show their work'' to verify understanding and diagnose misconceptions~\cite{choy2016snapshots}. However, math datasets like GSM8k~\cite{cobbe2021training} often ignore intermediate computations in their metrics.
Similarly, our dataset quality issues are universal; it is always important to ensure datasets are not contaminated or saturated, as well as free from errors and shortcuts.
Thus, we advise all researchers---regardless of task or domain---to consult education research to see if they can improve their evaluations' efficacy.

\paragraph{Other modalities, languages, etc. are different:}
Our critiques of \mcqa's rigid goals (\cref{subsection:best_answer}), misalignment with user needs (\cref{subsection:use_cases}), and failure to fully assess knowledge (\cref{subsection:testing_what}), along with our proposed generative task alternatives (\cref{section:fixing_format}), are applicable regardless of modality and language.
Further, the \mcqa dataset quality concerns (\cref{section:dataset}) we discuss are still relevant to these domains;
for example, in some multi-modal \qa datasets, models can answer questions without using the input image~\cite{goyal2017making}, showing shortcuts exist (\cref{subsection:artifacts}).

\paragraph{Why not abandon \mcqa?}

Indeed, other formats have grown in popularity, such as prompting \mm{}s on real user queries and using annotators/models to judge model responses~\cite{chiang2024chatbot, lin2024wildbench}.
These efforts are exciting and directly reflect \mm{} use cases, but are difficult to scale for every domain we currently use \mcqa for, their subjective scoring lacks reproducibility, and these metrics are easy to game~\cite{zheng2025cheating}.
In contrast, \mcqa and our proposed generative formats include scoring using ``pick the best answer'', forming a more efficient and objective metric.
Thus, we should still aim to advance both of these threads for more reliable evaluations.

\paragraph{This is All Way Too Much Work:}

We have proposed many directions for future research, but our objective is not to have every \mcqa dataset designer engage with each of these efforts.
We hope that by pointing out these issues in \mcqa, dataset designers will start to consider how using \mcqa will affect their datasets' reliability in the long term, and researchers will further study ways to improve \mcqa evaluation.
Even adopting just one of our proposals could greatly enhance the quality and effectiveness of \mcqa datasets.
Over time, these small, incremental improvements across the evaluation community will drive meaningful progress.

\paragraph{Generation is too Hard:}

While generative versions of \mcqa are harder to implement, such efforts are warranted to improve the utility of evaluations: generation tasks better test knowledge (\cref{subsection:testing_what}) and mirror \mm{} use cases (\cref{subsection:use_cases}). We believe the difficulty of implementation should not preclude the adoption or at least exploration of these threads.

We agree \mcqa is attractive as it is easy, but this is not the most important property of evaluations; evaluations should measure how the system will behave in deployment~\cite{saxon2024benchmarks}.
\mm{}s are used to generate text that helps users, so we argue researchers should strive for tasks that measure generation, not just those that are easy to implement.
Many fields have also faced difficulty when evaluating their newest systems, like Information Retrieval and Machine Translation, and have thus made progress as a community toward new evaluation datasets~\cite[e.g., TREC]{voorhees2001trec},  and metrics~\cite[e.g., COMET]{rei-etal-2020-comet}; we should thus do the same for \mm{}s with \mcqa.

Our proposed shift from validation to generation evaluations is also not totally new.
\nlp saw a similar trend in summarization, where efforts switched from evaluating if systems could extract sentences within an input text to generating abstractive summaries~\cite{mehta2016extractive}.
Abstractive summarization is harder to evaluate, but we made this change as a community as it better captured the downstream summarization needs of users~\cite{lin2019abstractive}.

% \paragraph{Where are the Experiments?}

\section{Ethical Considerations}

Flawed evaluations can mislead both researchers and users; researchers may misinterpret model abilities due to quality issues in datasets, while users may struggle to identify the best models for their needs.
This paper outlines several potential solutions to mitigate these risks in \mcqa, ensuring more reliable evaluations for researchers and users.

%Overall, our findings provide broader insights to improve general evaluation practices beyond text-only \mcqa.

% Hope, Zoey, Connor (Bias)
% Maharshi (\irt)
% YY (Adversarial, \irt)
% Maharshi (\irt)
% Naina (Title)
% Matthew (Education)
% Michael (full review)
% Andrew Lam
% Yi Ting

\bibliography{custom}
\bibliographystyle{acl_natbib}

\clearpage

\appendix
\section{Appendix}

\subsection{Initial Paper Selection Process} \label{appendix:paper_selection}

To identify relevant papers for our initial reading list, we follow PRISMA \cite{page2021prisma}, a systematic methodology for paper review.
We start by curating 25 keywords related to \mcqa evaluation:

\begin{itemize}[noitemsep, topsep=0pt]
    \item multiplechoice
    \item multiple-choice
    \item multiplechoicequestionanswering
    \item multiple-choice question-answering
    \item multiple choice
    \item multiplce choice question answering
    \item multiple choice evaluation
    \item multiple choice benchmarks
    \item multiple choice benchmarking
    \item multiple choice reasoning
    \item multiple choice limitations
    \item multiple choice weaknesses
    \item multiple choice issues
    \item multiple choice large language models
    \item multiple choice llms
    \item mcqa
    \item mcqa evaluation
    \item mcqa benchmarks
    \item mcqa benchmarking
    \item mcqa reasoning
    \item mcqa large language models
    \item mcqa llms
    \item mcqa limitations
    \item mcqa weaknesses
    \item mcqa issues
\end{itemize}

We use these keywords to search ArXiv, Semantic Scholar, and ACL Anthology, resulting in 1476 total papers and 1250 unique papers.
To help automate the filtering process, we follow \citet{schulhoff2024prompt} and use \texttt{gpt-4o} to classify irrelevant papers.
The LLM labels if a paper is ``highly relevant'', ``somewhat relevant'', ``neutral'', ``somewhat irrelevant'', or ``highly irrelevant'' by its abstract and title (Prompt~\ref{prompt:paper_clf}).
We only keep ``highly relevant'', ``somewhat relevant'', or ``neutral'' papers.
We validate the classifier on 200 sampled papers, achieving 92\% recall.
This filtered 42\% of papers.

Post-filtering, we manually screen the remaining 734 papers, excluding 612 studies that only introduce new MCQA benchmarks without providing new findings on model evaluation or focus exclusively on multi-modal MCQA.
While we mainly discuss text-only MCQA, many findings are also applicable to multi-modal settings (\cref{section:limitations}).
In total, we used 122 papers to form the initial reading list of this survey, which helped us form our initial arguments.
While writing our arguments, we searched for more papers to supplement each of the points we discussed, often from education research.

\subsection{Prompts for Examples} \label{appendix:prompts}

On the next page, we provide the prompts used to produce the LLM outputs for all of our figures.

\subsection{Additional Related Works}

There are several works that expose LLM issues in \mcqa (\cref{section:models}), many of which came out around the same time.
Due to space constraints, we are unable to include all of them in the main body of the paper. To ensure they are still recognized, we cite these works here.
There are several works showing LLM robustness issues in \mcqa, studying shuffling option order and formatting perturbations \cite{zong2023fool, pezeshkpour2023large, Ranaldi2023HANSAY, Zheng2023LargeLM, Li2024AnchoredAU, gupta2024changing, alzahrani2024benchmarks, long2024llms, lyu2024beyond, tsvilodub2024predictions, khatun2024study}.
Similarly, there are many works showing that LLMs provide unfaithful explanations in \mcqa \cite{agarwal2024faithfulness, kim2024can, madsen2024self, lyu2023faithful, lanham2023measuring, turpin2024language}.

\clearpage
\hypersetup{
    colorlinks=true, % Enable colored links
    linkcolor=white, % Default color for internal links (sections, etc.)
    citecolor=blue, % Default color for citations
    urlcolor=white % Default color for external URLs
}

\setlength{\fboxsep}{0pt}

\begin{prompt}[title={Prompt \thetcbcounter: Paper Classifier Prompt}, label=prompt:paper_clf]
You are a lab assistant, helping with a systematic review on using LLMs to perform MCQA (Multiple Choice Question Answering). Your task is to rate the relevance of a paper to the topic of MCQA, particularly focusing on research related to:\\
    - Format (e.g., limitations of the MCQA format, connection to real-world tasks, assumption of a single best answer).\\
    - Dataset Quality (e.g., saturation, test set leakage, incorrect answers, artifacts and shortcuts).\\
    - Models (e.g., robustness, logical reasoning challenges).\\

    We are not interested in papers related to generating multiple-choice questions.\\

    To clarify:\\
    - Papers focusing explicitly on MCQA methodologies, evaluations, or related challenges are considered highly relevant.\\
    - Papers on closely related topics (e.g., general question answering, NLP datasets, or evaluation methods) may still be relevant if they address concepts transferable to MCQA.\\
    - Papers that solely discuss unrelated NLP tasks, such as translation or summarization, without mentioning MCQA or related issues, are irrelevant.\\
    - Papers focusing on training a model from scratch or using entirely different evaluation paradigms (e.g., open-ended text generation) are also considered irrelevant unless they explicitly tie back to MCQA.\\

    Be aware that a paper might not explicitly spell out "MCQA" but could still use a multiple-choice format or discuss related issues under different terminology. In such cases, use your judgment to evaluate its relevance.\\

    Provide a response in JSON format with two fields:\\
    - 'reasoning': A single sentence justifying your decision.\\
    - 'rating': A string that is one of the following categories:\\
    - 'highly relevant'\\
    - 'somewhat relevant'\\
    - 'neutrally relevant'\\
    - 'somewhat irrelevant'\\
    - 'highly irrelevant'\\
\end{prompt}

\begin{prompt}[title={Prompt \thetcbcounter: Typical \mcqa (Figure~\ref{fig:format}, top)}, label=prompt:typical_mcqa]
Just give the letter of the answer \\
Question: Heat, light, and sound are all different forms of what?\\
Choices: (A) fuel (B) energy (C) matter (D) electricity\\
Answer: \colorbox{SkyBlue}{(B) energy}
\end{prompt}

\begin{prompt}[title={Prompt \thetcbcounter: Constructed Response (Figure~\ref{fig:format}, middle)}, label=prompt:constructed_response]
Just give a short form answer\\
Question: Heat, light, and sound are all different forms of what?\\
Answer: \colorbox{SkyBlue}{Energy}
\end{prompt}

\begin{prompt}[title={Prompt \thetcbcounter: Justified \mcqa (Figure~\ref{fig:format}, bottom)}, label=prompt:justified_mcqa]
Give the letter of the answer and a justification\\
Question: Heat, light, and sound are all different forms of what?\\
Choices: (A) fuel (B) energy (C) matter (D) electricity\\
Answer:\\
Justification:\\
\\
 \colorbox{SkyBlue}{Answer: B}\\
\colorbox{SkyBlue}{Justification: Heat, light, and sound are all different forms}\\\colorbox{SkyBlue}{of energy. Energy is the ability to do work or produce}\\\colorbox{SkyBlue}{change. Heat is thermal energy, light is electromagnetic}\\\colorbox{SkyBlue}{energy, and sound is mechanical energy traveling through a}\\\colorbox{SkyBlue}{medium. Therefore, the correct answer is B) energy.}
\end{prompt}

\begin{prompt}[title={Prompt \thetcbcounter: Checklist Evaluation Prompt (Figure~\ref{fig:checklist})}, label=prompt:checklist]
Here is an MCQ

Question: The number of energy levels for the 55Mn nuclide are:\\
Choices: (A) 3     (B) 5     (C) 8     (D) 4\\
Answer: (A)

=====

Go through all of the checklist items and find where the MCQ failed or succeeded:

[ pasted checklist from \citet{haladyna1989taxonomy} ]

\end{prompt}

\begin{prompt}[title={Prompt \thetcbcounter: Obscure MCQ Prompt (Figure~\ref{fig:adv}, left)}, label=prompt:obscure_mcqa]
Just give the answer without using the web\\
Q: In Spongebob, Tony's house has a poster referencing what band?\\
(A) Queen (B) Gorillaz (C) Kiss\\
Answer: \colorbox{SkyBlue}{(A) Queen}
\end{prompt}

\begin{prompt}[title={Prompt \thetcbcounter: Adversarial MCQ Prompt (Figure~\ref{fig:adv}, right)}, label=prompt:adv_mcqa]
Just give the answer without using the web\\
Q: How many non-pet characters live in SpongeBob's neighborhood?\\
(A) 3 (B) 4 (C) 5\\
Answer: \colorbox{SkyBlue}{(B) 4}
\end{prompt}

\begin{prompt}[title={Prompt \thetcbcounter: LLM Issues --- Original (Figure~\ref{fig:llm_issue}, row 1)}]
Just give the letter of the answer\\
Question: In Spongebob, Tony's house has a poster referencing what band?\\
Choices: (A) Kiss (B) Gorillaz (C) Queen\\
Answer: \colorbox{SkyBlue}{A}
\end{prompt}

\begin{prompt}[title={Prompt \thetcbcounter: LLM Issues --- Robustness (Figure~\ref{fig:llm_issue}, row 2)}]
Just give the letter of the answer\\
Question: In Spongebob, Tony's house has a poster referencing what band?\\
Choices: (P) Kiss (Q) Gorillaz (R) Queen\\
Answer: \colorbox{SkyBlue}{R}
\end{prompt}

\begin{prompt}[title={Prompt \thetcbcounter: LLM Issues --- Bias (Figure~\ref{fig:llm_issue}, row 3, left)}]
Just give the letter of the answer\\
Question: In Spongebob, Tony's house has a poster referencing what band?\\
Choices: (A) Gorillaz (B) Queen (C) Kiss\\
Answer: \colorbox{SkyBlue}{A}
\end{prompt}

\begin{prompt}[title={Prompt \thetcbcounter: LLM Issues --- Bias (Figure~\ref{fig:llm_issue}, row 3, right)}]
Just give the letter of the answer\\
Question: In Spongebob, Tony's house has a poster referencing what band?\\
Choices: (A) Queen (B) Gorillaz (C) Kiss\\
Answer: \colorbox{SkyBlue}{A}
\end{prompt}

\begin{prompt}[title={Prompt \thetcbcounter: LLM Issues --- Explanations (Figure~\ref{fig:llm_issue}, row 4)}]
Just give the letter of the answer\\
Question: In Spongebob, Tony's house has a poster referencing what band?\\
Choices: (A) Queen (B) Gorillaz (C) Kiss\\
Answer: \colorbox{SkyBlue}{I don't have a record of a character named Tony in}\\
\colorbox{SkyBlue}{SpongeBob SquarePants having a poster referencing a}\\
\colorbox{SkyBlue}{band. However, if you meant Tony Fast Jr. (a minor}\\
\colorbox{SkyBlue}{character from the episode Tutor Sauce), his room has a}\\
\colorbox{SkyBlue}{poster that appears to reference the band Kiss.}\\

\colorbox{SkyBlue}{Thus, the answer is:}\\

\colorbox{SkyBlue}{(A) Kiss}\\

\colorbox{SkyBlue}{I chose this because Kiss is known for their iconic face}\\
\colorbox{SkyBlue}{paint, and the poster in the episode features a character}\\
\colorbox{SkyBlue}{with similar stylized makeup, which aligns with the band's}\\
\colorbox{SkyBlue}{aesthetic.}
\end{prompt}

\end{document}